\documentclass[accepted]{uai2021} 
\pdfoutput=1

\usepackage[american]{babel}

\usepackage{natbib} 
\usepackage{mathtools} 
\usepackage{booktabs} 
\usepackage{tikz} 

\usepackage{authblk}

\title{LGD-GCN: Local and Global Disentangled Graph Convolutional Networks}

\author[1, 2]{Jingwei Guo}
\author[2]{Kaizhu Huang}
\author[1]{Xinping Yi}
\author[2]{Rui Zhang}
\affil[1]{
University of Liverpool, Liverpool, UK
}
\affil[2]{
    Xi'an Jiaotong-Liverpool University, Suzhou, China
}
\affil[ ]{Jingwei.Guo@liverpool.ac.uk, Kaizhu.Huang@xjtlu.edu.cn}
\affil[ ]{Xinping.Yi@liverpool.ac.uk, Rui.Zhang02@xjtlu.edu.cn}

\usepackage[ruled,vlined]{algorithm2e}
\usepackage{algpseudocode}
\usepackage{amssymb}
\usepackage{amsthm}
\usepackage{amsmath}
\usepackage{subcaption}

\usepackage{graphicx}
\usepackage{multirow}

\usepackage{accents}
\usepackage{enumitem}
\usepackage{hyperref}

\begin{document}
\maketitle

\begin{abstract}
Disentangled Graph Convolutional Network (DisenGCN) is an encouraging framework to disentangle the latent factors arising in a real-world graph. However, it relies on disentangling information heavily from a local range (i.e., a node and its 1-hop neighbors), while the local information in many cases can be uneven and incomplete, hindering the interpretabiliy power and model performance of DisenGCN. In this paper\footnote{This paper is a lighter version of \href{https://jingweio.github.io/assets/pdf/tnnls22.pdf}{"Learning Disentangled Graph Convolutional Networks Locally and Globally"} where the results and analysis have been reworked substantially. Digital Object Identifier \url{https://doi.org/10.1109/TNNLS.2022.3195336}.}, we introduce a novel Local and Global Disentangled Graph Convolutional Network (LGD-GCN) to capture both local and global information for graph disentanglement. LGD-GCN performs a statistical mixture modeling to derive a factor-aware latent continuous space, and then constructs different structures w.r.t. different factors from the revealed space. In this way, the global factor-specific information can be efficiently and selectively encoded via a message passing along these built structures, strengthening the intra-factor consistency. We also propose a novel diversity promoting regularizer employed with the latent space modeling, to encourage inter-factor diversity. Evaluations of the proposed LGD-GCN on the synthetic and real-world datasets show a better interpretability and improved performance in node classification over the existing competitive models. Code is available at \url{https://github.com/jingweio/LGD-GCN}.
\end{abstract}

\section{Introduction}
Graphs are emerging as an insightful structured data modeling technique for capturing the similarity between data samples and the relationship between entities. To mine the domain-specific knowledge in graph structured data, Graph Convolutional Networks~(GCNs) have been proposed to integrate the topological patterns and content features~\citep{Kipf2017SemiSupervisedCW}, demonstrating excellent expressive power and growing popularity in various graph learning tasks, such as node classification, link prediction, and recommendation~\citep{Wu2020ACS,chen2020handling}.

Most state-of-the-art methods, such as~\citep{Kipf2017SemiSupervisedCW,Hamilton2017InductiveRL,Velickovic2018GraphAN}, study node representations in a holistic approach, i.e., they interpret the node neighborhood as a whole without considering the within-distinctions. By contrast, a real-world graph typically contains multiple heterogeneous node relations which in many cases are implicitly determined by various latent factors shaping node aspects. For instance, a user in a social network, usually links with different persons for different reasons, such as family, work, and/or hobby, which potentially characterize the user from different perspectives. The existing holistic approaches usually fail to disentangle these latent factors, rendering the learned representations hardly explained and less informative.

Recently, Disentangled Graph Convolutional Network~(DisenGCN)~\citep{Ma2019DisentangledGC} offers a promising framework to disentangle the latent factors behind graph data via a neighborhood partition. Despite the novel design, DisenGCN heavily relies on local node neighborhood, which may bring unexpected issues. First, the information from local ranges can be significantly varied across the entire graph. Solely depending on it, DisenGCN could easily produce latent representations losing consistent meaning of the associated factor. That may weaken the intra-factor correlation between disentangled features and leads to diminished interpretability. Second, the local neighborhood information can be scarce and limited especially considering sparse graphs, prohibiting DisenGCN from generating informative node aspects and yielding favourable performance boost. A detailed discussion can be seen later in Section~\ref{sec:limit}.

 
To tackle this limitation, in this paper, we propose a novel framework, termed as Local and Global Disentangled Graph Convolutional Network (LGD-GCN), to learn disentangled node representations capturing \emph{both local and global} graph information. The central idea is that disentangling the latent factors inherent in a graph can benefit from a latent continuous space which uncovers the underlying factor-aware node relations. Specifically, we first leverage the neighborhood routing mechanism to locally disentangle node representations into multiple latent units pertinent to different factors. Then, we propose to guide the disentanglement from a global perspective.

\begin{figure}[t]
     \centering
     \begin{subfigure}[b]{0.235\textwidth}
         \centering
         \includegraphics[width=1\columnwidth]{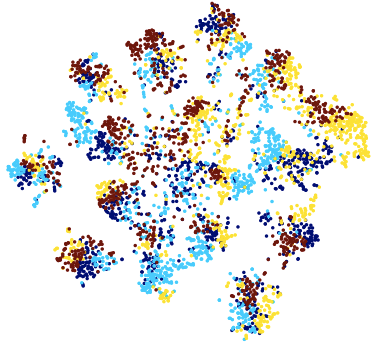}
         \caption{\small{DisenGCN}}
         \label{fig:disengcn_bad}
     \end{subfigure}
     \hfill
     \begin{subfigure}[b]{0.235\textwidth}
         \centering
         \includegraphics[width=1\columnwidth]{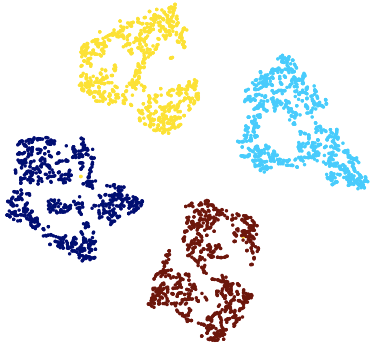}
         \caption{\small{LGD-GCN (Ours)}}
         \label{fig:lgd_good}
     \end{subfigure}
\caption{Visualization of the disentangled latent units w.r.t. four latent factors on a synthetic graph. Points with a different color mean the disentangled latent units (for all nodes) of a different latent factor. In sharp contrast to DisenGCN, our LGD-GCN displays a highly disentangled pattern with strong intra-factor consistency and inter-factor diversity; it indicates high (low) intra-factor (inter-factor) correlations between features}
\label{fig:DisentangleSensi}
\end{figure}

To this end, our approach performs a mixture statistical modeling over the locally disentangled latent units, to derive a factor-aware latent continuous space. This enables a different component or mode, specific to a latent factor, in a different region of the latent space~\citep{Ghahramani1996TheEA}. After that, we manage to build a different structure by connecting near neighbors in a different region of the revealed latent space. These latent structures disclose the underlying factor-aware relations between nodes. Employing message passing along them can efficiently and selectively encode the global factor-specific information, which enhances \emph{intra-factor consistency}, i.e., the consistent meaning of disentangled latent units  w.r.t. the associated factor. Furthermore, we also design a novel diversity promoting regularizer to encourage \emph{inter-factor diversity}. Practically, it enforces the disentangled latent units related to different factors to fall into separate clusters in the latent space so as to enhance the disentangled informativeness. In sharp contrast to DisenGCN, Fig.~\ref{fig:DisentangleSensi} clearly visualizes the benefit of learning disentangled node representations \emph{both locally and globally}. Our contributions are summarized as below:
\begin{itemize}[noitemsep,nolistsep]
    \item We argue that DisenGCN may bring unexpected issues by heavily relying on local graph information. Empirical analysis shows that DisenGCN may learn latent representations with weakly disentangled factors, and especially its boost performance becomes minor while facing sparse graphs.
    \item We propose a novel Local and Global Disentangled framework for Graph Convolutional Networks (LGD-GCN) to infer the latent factors underlying the graph data. Incorporating both local and global information, LGD-GCN can disentangle node representations with enhanced intra-factor consistency and promoted inter-factor diversity.
    \item Extensive evaluations on synthetic and real-world datasets demonstrate that LGD-GCN provides a better interpretability and improved performance in node classification compared to other state-of-the-arts.
\end{itemize}

\section{Background and motivation}
\subsection{Conventional GNNs}
Graph neural networks (GNNs) are powerful machine learning models in dealing with graph-structured data, where the input data are modeled as graphs. A graph is denoted as $G = (V, E)$ with $V$ being the set of nodes and $E$ being the set of edges. Given two distinct nodes $u, v \in V$ with $u \neq v$, we define $(u, v) \in E$ if $u$ and $v$ are connected with an edge, and the neighborhood of node $u$ as $N_u = \{v|(u,v) \in E\}$. For attributed graphs, each node $u$ usually has an initial representation $\mathbf{h}_u^{(0)} \in \mathbb{R}^{d_{0}}$. GNNs are used to mine the underlying relationship between nodes according to their attributes $\{\mathbf{h}_u^{(0)}| \forall u \in V \}$ for (semi-)supervised learning tasks such as node classification and link prediction.

In the past years, an increasing number of GNN models have been proposed~\citep{Wu2020ACS}. Most of them can be generalized by a message passing mechanism~\citep{gilmer2017neural}, where the node attributes are exchanged through the graph edges following a neighborhood aggregation strategy descibed below.
\begin{equation*}
\mathbf{h}_u \leftarrow \mathbf{UPDATE}(\mathbf{AGGREGATE}(\{\mathbf{h}_v| \forall v \in N_u\}),\mathbf{h}_u),
\end{equation*}
where the $\mathbf{AGGREGATE}$ operation is to aggregate information from a node neighborhood, and the $\mathbf{UPDATE}$ operation is to combine these information to update the node's attributes. Such a strategy works in an iterative way to learn node representations. For graph-level representations, a readout operation, such as a simple mean or sum, can be applied to summarize the overall information.




\subsection{disentangled node representation}

Albeit promising several learning tasks, most GNNs treat the neighborhood as a whole and ignore the inner-differences, learning noninterpretable representations. To address this issue, DisenGCN~\citep{Ma2019DisentangledGC} was proposed by hypothesizing nodes are connected due mainly to different kinds of relationship; and there are $M$ inherent factors determining edge connections and potentially shaping nodes from $M$ aspects.

DisenGCN aims to disentangle each node representation into multiple latent units w.r.t. different latent factors. In each layer, given a node $u$ and its neighborhood $N_u$, the node representations, $\{\mathbf{h}_i | \forall i \in \{u\} \cup N_u\}$, will be first projected onto $M$ subspaces using different channels. In each channel $m$ ($m = 1,2,...,M$), the projected representation of node $i$ is given by
\begin{equation}
\mathbf{z}_{i, m} = \frac{\sigma(\mathbf{W}_m^T\:\mathbf{h}_i + \mathbf{b}_m)}{\|\sigma(\mathbf{W}_m^T\:\mathbf{h}_i + \mathbf{b}_m)\|_2}
\label{eq:channels_mp}
\end{equation}
where $\mathbf{W}_m \in \mathbb{R}^{d_{in} \times \frac{d_{out}}{M}}$ and  $\mathbf{b}_m \in \mathbb{R}^\frac{d_{out}}{M}$ are learnable parameters, and $\sigma$ is an activation function.
A neighborhood routing mechanism, detailed in~\citep[Algorithm-1]{Ma2019DisentangledGC}, then iteratively partitions all the neighbors into different clusters. After that, independent information aggregations are applied over them in different channels, to attain the disentangled latent units for node $u$, $\{\hat{\mathbf{z}}_{u, m} \in \mathbb{R}^{\frac{d_{out}}{M}}| \forall m = 1,2,...,M\}$. Finally, the disentangled node representation is obtained by concatenation, $\hat{\mathbf{h}}_u = \hat{\mathbf{z}}_{u, 1} \oplus \hat{\mathbf{z}}_{u, 2} \oplus ... \oplus \hat{\mathbf{z}}_{u, M}$ and $\hat{\mathbf{h}}_u \in \mathbb{R}^{d_{out}}$.

\subsection{Limitations of DisenGCN}\label{sec:limit}
While DisenGCN reveals certain latent factors, we argue that it tends to produce weakly disentangled representations and yield limited performance boost because of its heavy reliance on local graph information. To validate our argument, we further provide two experimental investigations over the graph synthesized with four latent factors (see details in Section~\ref{sec:exp_set}).

First, we visualize the disentangled latent units of DisenGCN using t-SNE in Fig.~\ref{fig:disengcn_bad}. At the micro-level, we can observe the separability between points with different colors in some regions. But, when it comes to the macro-level, all points unexpectedly fall into discrete clusters and mixed together, indicating a weak disentanglement. This is because the disentangled latent units by DisenGCN may preserve some specific micro-meanings of the factor, but losing the consistent macro-meaning (\emph{intra-factor consistency}). Additionally, DisenGCN only considers disentangling representations in different channels without ensuring the diversity between those w.r.t. different factors (\emph{inter-factor diversity}). The learned representations thereby are prone to preserve the redundant information, partially explaining Fig.~\ref{fig:disengcn_bad}.

Second, we further augment the synthetic graph by tuning the $p$ value (it controls the density of the synthetic graph as described in Section~\ref{sec:exp_set}) to generate graphs with multiple average neighborhood sizes. We then apply GCN~\citep{Kipf2017SemiSupervisedCW} and DisenGCN to train for multi-label classification, and report the F1 scores in Table~\ref{tab:disengcn_limitation}. From the table, the relative improvements reduce from approximately 5\% to 1\% as the average neighborhood size decreasing from 40 to 6. The result meets the expectation. DisenGCN may perform well on a dense graph by learning disentangled representations. However, sparsing the input graph (limiting the accessible local information) can negatively affect the performance boost, which reflects the heavy local reliance of DisenGCN.


\begin{table}[t]
\centering
\setlength{\abovecaptionskip}{4pt}\makeatletter\def\@captype{table}\makeatother
\caption{\small{Micro (Top) and Macro (Bottom) F1 scores (\%) on graphs synthesized with four latent factors but different average neighborhood sizes}}
\resizebox{0.485\textwidth}{!}{
\begin{tabular}{cccccc}
\hline
\multirow{2}{*}{\textbf{Methods}}  & \multicolumn{5}{c}{\textbf{Average Neighborhood Sizes}} \\ \cline{2-6}
& 40        & 30        & 20        & 10       & 6       \\ \hline
GCN                                
&79.5$\pm$0.8   &75.5$\pm$0.7   &66.1$\pm$0.9   &47.2$\pm$0.6   &37.2$\pm$0.9         \\
DisenGCN                          
&84.1$\pm$1.0   &79.5$\pm$0.7   &69.0$\pm$1.0   &48.8$\pm$0.9   &38.4$\pm$0.8         \\
\textbf{Improvements}                      
&\textbf{+4.6\%}    &\textbf{+4.0\%}    &\textbf{+2.9\%}    &\textbf{+1.6\%}    &\textbf{+1.2\%}
\\ \hline
GCN                                        
&78.3$\pm$0.9   &75.0$\pm$0.8   &65.8$\pm$1.0   &45.8$\pm$0.6   &36.7$\pm$0.8         \\
DisenGCN                          
&82.9$\pm$1.1   &78.9$\pm$0.7   &68.3$\pm$1.0   &47.4$\pm$1.0   &37.7$\pm$0.8         \\
\textbf{Improvements}                      
&\textbf{+4.6\%}            &\textbf{+3.9\%}              &\textbf{+2.5\%}            &\textbf{+1.6\%}            &\textbf{+1.0\%}         \\ \hline
\end{tabular}}
\label{tab:disengcn_limitation}
\end{table}

\begin{figure*}[t]
\centering
\includegraphics[width=1.0\textwidth]{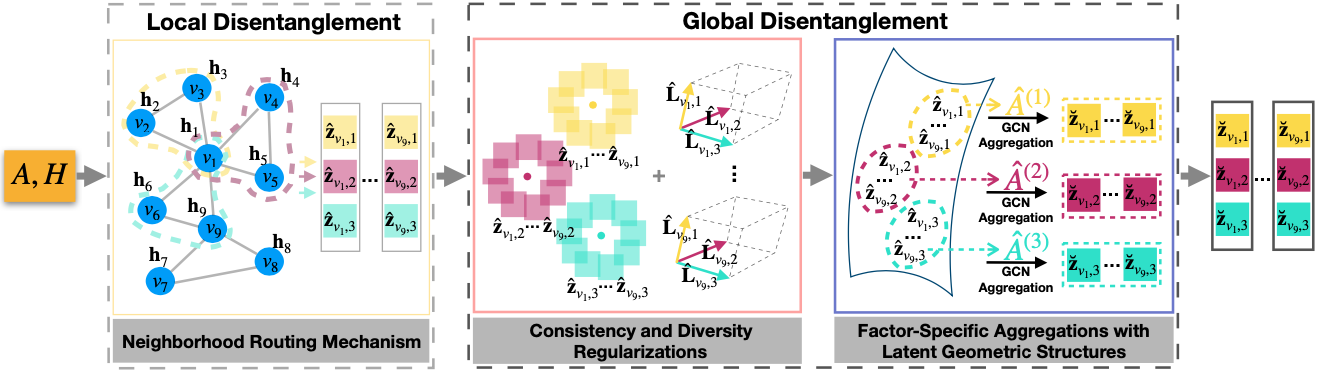}
\caption{\small{Illustrative example of the LGD-GCN layer with $M=3$ latent factors. First, the node representations are locally disentangled by leveraging the neighborhood routing mechanism. These disentangled representations are then modeled in a latent continuous space, and promoted with consistent and diverse latent factors globally, from which geometric structures are constructed for further aggregation.}}
\label{fig:lgd_layer}
\end{figure*}

\section{Local and Global Disentangled GCN}
We present an novel method for Graph Convolutional Networks (LGD-GCN) to disentangle node representations both locally and globally, as presented in Fig.~\ref{fig:lgd_layer}. By hiring the neighborhood routing mechanism~\citep{Ma2019DisentangledGC}, we first attain disentangled latent units preserving local graph information w.r.t. different latent factors. However, these disentangled units are prone to be weakly disentangled without incorporating global information and being properly regularized. In the following, we show how to enhance the disentanglement from a global perspective, via disclosing the underling factor-aware relations between nodes, to learn a better disentangled representations with strengthened intra-factor consistency and promoted inter-factor diversity.

\subsection{Modeling Latent Continuous Space}
Extending the hypothesis in DisenGCN from local neighborhood to global graph, we assume that the locally disentangled units for all nodes, $\{\hat{\mathbf{z}}_{i, m} \in \mathbb{R}^{\frac{d_{out}}{M}}| \forall i \in V, \forall m = 1,2,...,M\}$, are generated from a gaussian mixture distribution with equal mixture weights, s.t.
\begin{equation} \label{eq:gmd}
p(\hat{\mathbf{z}}_{i,m}) = \frac{1}{M}\sum^M_{e=1}\mathcal{N}(\hat{\mathbf{z}}_{i,m};\boldsymbol{\mu}_e, \boldsymbol{\Sigma}_e),
\end{equation}
where $\boldsymbol{\mu}_e \in \mathbb{R}^{\frac{d_{out}}{M}}$ and $\boldsymbol{\Sigma}_e \in \mathbb{R}^{\frac{d_{out}}{M}\times \frac{d_{out}}{M}}$ are the mean and the covariance associated with latent factor $e$. Then, we employ this assumption to learn factor-specific means and covariances to regularize the disentangling of the latent units. Specifically, we maximize the conditional likelihood of the latent units $\hat{\mathbf{z}}_{i,m}$ (for each node $i$ and each factor $m$) w.r.t. the associated factor $m$. It is equivalent to minimizing the negative log term expressed in Eq.~(\ref{eq:SPACE}) after removing constants.
\begin{equation} \label{eq:SPACE}
\mathcal{L}_{i,m} = (\hat{\mathbf{z}}_{i,m}-\boldsymbol{\mu}_m)^T\boldsymbol{\Sigma}_m^{-1}(\hat{\mathbf{z}}_{i,m}-\boldsymbol{\mu}_m)
\end{equation}

Minimizing the term $\mathcal{L}_{i,m}$ is equivalent to minimizing the Mahalanobis Distance~\citep{de2000mahalanobis} between the disentangled latent unit and its globally inferred center. It derives a latent continuous space where the disentangled latent units are encouraged to be more discriminative with respect to their centers, and to carry the type of global factor-specific information shared by all nodes.

\subsection{Constructing Latent Structures} \label{sec:geo_struc}
Although node relations are naturally presented in graph data, we believe that they are not always optimal for disentangled graph learning. Taking a huge and sparse graph as an example. It is difficult for most nodes to absorb sufficient information, coming from a small number of their neighbors (one or two in most cases), to learn disentangled representations w.r.t. latent factors in a larger number. On the other hand, the raw graph may not contain the desired topologies after projecting node features in different channels. As such, disclosing the underlying factor-aware relations between nodes from the disentangled latent space becomes a promising alternative.

The previously modeled latent space enables a different component or mode, specific to a different latent factor, in a different region. It would be reasonable to apply a proper graph construction algorithm over different regions to obtain latent structures specific to different factors. We expect these built structures uncovering the factor-aware relations between nodes, and selecting a sufficient number of latent neighbors (from the entire graph) for each node in shaping node aspects. Accordingly, the global factor-specific information can be efficiently and selectively encoded, by employing a simple message passing scheme independently along these different structures, to strengthen the \emph{intra-factor consistency}.

Here, we list two popular methods for building graphs from data using local neighborhood in latent space:

1) \textbf{k-Nearest-Neighbor (kNN)}: It connects every point to its k$^{th}$ nearest neighbors, given a pairwise distance $d(\mathbf{z}_i,\mathbf{z}_j)$. Formally, the adjacency matrix $\mathbf{A}^{kNN} \in \{0,1\}^{N \times N}$ is defined as:
\begin{equation*} \label{eq:cknn}
\small \mathbf{A}^{knn}_{i, j} = \begin{cases}
1 & d(\mathbf{z}_i,\mathbf{z}_j) \leq d(\mathbf{z}_i,\mathbf{z}_j^{(k)})\ or\ d(\mathbf{z}_i,\mathbf{z}_j) \leq d(\mathbf{z}_i^{(k)},\mathbf{z}_j)\\
0 & \text{otherwise}
\end{cases}
\end{equation*}
where $\mathbf{z}_i^{(k)}$ and $\mathbf{z}_j^{(k)}$ denote the k$^{th}$ nearest neighbors of $\mathbf{z}_i$ and $\mathbf{z}_j$, respectively.

2) \textbf{Continuous k-Nearest-Neighbor (CkNN)}~\citep{berry2016consistent}: It is a discrete version of kNN for removing kNN's sensitivity to the density parameter $k$. Similary, the adjacency matrix $\mathbf{A}^{CkNN} \in \{0,1\}^{N \times N}$ is defined as:
\begin{equation*} \label{eq:cknn}
\small \mathbf{A}^{cknn}_{i, j} = \begin{cases}
1 & d(\mathbf{z}_i,\mathbf{z}_j) < \sqrt{d(\mathbf{z}_i,\mathbf{z}_i^{(k)})d(\mathbf{z}_j,\mathbf{z}_j^{(k)})} \\
0 & \text{otherwise}
\end{cases}
\end{equation*}

In this paper, we apply the same message passing function in GCN~\citep{Kipf2017SemiSupervisedCW} to aggregate the factor-specific node information along these constructed structures, following Eq.~(\ref{eq:facagg}).
\begin{equation} \label{eq:facagg}
\breve{\mathbf{Z}}^{(m)} = {\breve{\mathbf{D}}^{(m)}}^{-\frac{1}{2}}\breve{\mathbf{A}}^{(m)}{\breve{\mathbf{D}}^{(m)}}^{-\frac{1}{2}}\hat{\mathbf{Z}}^{(m)}
\end{equation}
Here, $\hat{\mathbf{A}}^{(m)}$ refers to the built structures w.r.t. latent factor $m$ from $\{\hat{\mathbf{z}}_{i,m}| \forall i \in V\}$, $\breve{\mathbf{A}}^{(m)}=\hat{\mathbf{A}}^{(m)} + \mathbf{I}$, $\breve{\mathbf{D}}^{(m)}_{i,i}=\sum_j\breve{\mathbf{A}}^{(m)}_{i, j}$, $\breve{\mathbf{D}}^{(m)}_{i,j}=0$ in case of $i\neq j$, and $\hat{\mathbf{Z}}^{(m)}$ is the feature matrix with each column being $\hat{\mathbf{z}}_{i,m}$ for node $i$ in $V$. 
Particularly, we adopt the Euclidean distance as the pairwise distance $d(,)$, and denote this proposed module as $\mathbb{LG}_{agg}$.

\subsection{Promoting Inter-Factor Diversity}
Diversity-promoting learning aims to encourage different components in latent space models to be mutually uncorrelated and different, and has been widely
studied~\citep{Xie2018DiversityPromotingAL,Xie2016DiversityLT}. In the previous sections, we derived a factor-aware latent continuous space and built structures for encoding factor-specific node information from a global range, to strengthen the \emph{intra-factor consistency}. However, without being regularized to be different with respect to different latent factors, the disentangled latent units may preserve redundant information of other irrelevant latent factors.

In this paper, we propose to promote the \emph{inter-factor diversity} to capture the unique information in disentangled latent units. Particularly, we define the diversity on the conditional likelihoods (given different factors) of each disentangled latent unit in latent space. Inspired by the Determinant Point Process~\citep{Kulesza2012DeterminantalPP}, we formulate the disentanglement diversity for each node $i$ as
\begin{equation}
\mathbb{DD}_i = \det(\hat{\mathbf{F}}_i^{T}\hat{\mathbf{F}}_i) \label{eq:DD},
\end{equation}
where $\hat{\mathbf{F}}_i = <\hat{\mathbf{L}}_{i,1},...,\hat{\mathbf{L}}_{i,m},...,\hat{\mathbf{L}}_{i,M}>$, $\hat{\mathbf{L}}_{i,m}=\|\mathbf{L}_{i,m}\|_2$, and $\mathbf{L}_{i,m}=<\mathcal{N}(\hat{\mathbf{z}}_{i,m};\boldsymbol{\mu}_1, \boldsymbol{\Sigma}_1),...,\mathcal{N}(\hat{\mathbf{z}}_{i,m};\boldsymbol{\mu}_M, \boldsymbol{\Sigma}_M)>$ contains the conditional likelihoods (given $M$ factors) of the disentangled latent unit $\hat{\mathbf{z}}_{i,m}$.

By the property of Determinant~\citep{Bernstein2005MatrixMT}, $\mathbb{DD}_i$ is equal to the volume spanned by $\{\hat{\mathbf{L}}_{i,m}|\forall m = 1,2,...,M\}$, elegantly providing an intuitive geometric interpretation as shown in Fig.~\ref{fig:lgd_layer} with $M=3$. Maximizing $\mathbb{DD}_i$ encourages $\mathbf{L}_{i,1}, \mathbf{L}_{i,2}, ..., \mathbf{L}_{i,M}$ to be orthogonal to each other, i.e., enforcing the disentangled latent units to fall into separated regions of the statistical latent space; it essentially enhances the disentangled informativeness and promotes the \emph{inter-factor diversity}.


\subsection{Network Architecture}
In this section, we detail the general network architecture of the proposed LGD-GCN for performing node-level tasks. The pseudocode of a LGD-GCN's layer is presented in Algorithm~\ref{alg:lgd_gcn}, and it is desirable to stack multiple LGD-GCN's layers to sufficiently exploit the graph data. 

Specifically, we adopt ReLU activation function in Eq.~(\ref{eq:channels_mp}) and append a dropout layer~\citep{Srivastava2014DropoutAS} in the end of each LGD-GCN's layer which is only enabled in training. We can then have the output of layer $l$ as $\{\breve{\mathbf{h}}_i^{(l)} | \forall i \in V\} = \text{Dropout}(F^{(l)}(\{\breve{\mathbf{h}}_i^{(l-1)} | \forall i \in V\}))$,
where $1\leq l\leq L$, $\breve{\mathbf{h}}_i^{(0)}=\mathbf{h}_i^{(0)}$, $L$ denotes the number of stacked hidden layers, $F^{(l)}$ refers to LGD-GCN's $l^{th}$ layer. Finally, a fully connected layer is taken to map the learned node representations into another dimension, e.g. a class-level for node classification, expressed as $\mathbf{Y}_i^{(L+1)} = {\mathbf{W}^{(L+1)}}^T\breve{\mathbf{h}}_i^{(L)} + \mathbf{b}^{(L+1)}$, where ${\mathbf{W}^{(L+1)}} \in \mathbb{R}^{d_{out} \times C}$, $\mathbf{b}^{(L+1)} \in \mathbb{R}^{C}$, and $C$ is the number of class. 

In this work, we focus on the task of node classification. To incorporate Eq.~(\ref{eq:SPACE}) and Eq.~(\ref{eq:DD}) into the final optimization problem, we leverage them into two regularization terms for each node $i$, as expressed below.
\begin{equation} \label{eq:reg_losses}
\mathcal{L}_{space}^i = \frac{1}{M}\sum^{M}_{m=1}\mathcal{L}_{i,m},\  \mathcal{L}_{div}^i = -\log(\mathbb{DD}_i)
\end{equation}
For $\mathcal{L}_{i,m}$, we update $\mu_m$ and $\Sigma_m$ ($m=1,2,...,M$) in each layer iteratively with the newly computed values after each training epoch by an update rate $U_r$. To adaptively modify the influential power of these two regularization terms in different layers, we apply a layer loss weight, $\lambda^{(l)} = 10^{l-L}$. It makes the influence of the regularization terms grows bigger as the layer goes deeper within a proper range. Then, we can formulate the final loss in Eq.~(\ref{eq:ttloss}) with coefficients $\lambda_{space}$ and $\lambda_{div}$ for trade-off.
\begin{equation} \label{eq:ttloss}
    \mathcal{L}_{total} = \mathcal{L}_{cls} + \sum_{l=1}^L\lambda^{(l)}(\lambda_{space}\mathcal{L}_{space}^{(l)} + \lambda_{div}\mathcal{L}_{div}^{(l)})
\end{equation}
Here, for single-label node classification, $\mathcal{L}_{cls} = -\frac{1}{|V|}\sum_{i }^V\mathbb{Y}_i^T\log(\text{softmax}(\mathbf{Y}_i^{(L+1)}))$, and for multi-label classification, $\mathcal{L}_{cls} = -\frac{1}{|V|}\sum_{i }^V\mathbb{Y}_i^T\log(\text{sigmoid}(\mathbf{Y}_i^{(L+1)})) + (1-\mathbb{Y}_i)^T\log(1-\text{sigmoid}(\mathbf{Y}_i^{(L+1)}))$, given $\mathbb{Y}_i \in \mathbb{R}^C$ being the ground truth label of node $i$ in one hot encoding. The end-to-end optimization procedures are displayed in the supplemental material.

\begin{algorithm}[t]
\caption{\small{LGD-GCN's Layer}}\label{alg:lgd_gcn}
\SetAlgoLined
\small{
\textbf{Input}: $\{\mathbf{h}_i \in \mathbb{R}^{d_{in}}|\forall i \in V\}$; $M$: the number of latent factors; $T$: the routing iterations' number of the Neighborhood Routing Mechanism (NRM); \\
\textbf{Parameter}: $\mathbf{W}_m \in \mathbb{R}^{d_{in} \times \frac{d_{out}}{M}}, \mathbf{b}_m \in \mathbb{R}^{\frac{d_{out}}{M}}, \boldsymbol{\mu}_m \in \mathbb{R}^{\frac{d_{out}}{M}}, \boldsymbol{\Sigma}_m \in \mathbb{R}^{\frac{d_{out}}{M} \times \frac{d_{out}}{M}}$, $\forall m = 1,2,...,M$ \\
\For{$i \in V$}{
    $\mathbf{z}_{i, 1},\mathbf{z}_{i, 2},...,\mathbf{z}_{i, M} \gets \mathbf{h}_i$ by Eq.~(\ref{eq:channels_mp}). \\
}
\For{$i \in V$}{
    $\hat{\mathbf{z}}_{i,m} \gets \{\mathbf{z}_{i,m}\}\cup\{\mathbf{z}_{v,m}|\forall v \in N_i\}, \forall m = 1,2,...,M$ by NRM with $T$ routing iterations. \\
    Minimize $\mathcal{L}_{space}^i$ and $\mathcal{L}_{div}^i$ by Eq.~(\ref{eq:reg_losses}). \\
}
\For{$m = 1,2,...,M$}{
    Construct structure $\mathbf{G}^{(m)}$ with $\mathbf{A}^{(m)}$ from $\{\hat{\mathbf{z}}_{i,m} | \forall i \in V\}$. \\
    $\{\breve{\mathbf{z}}_{i,m} | \forall i \in V\} \gets \{\hat{\mathbf{z}}_{i,m} | \forall i \in V\}$ by Eq.~(\ref{eq:facagg}). \\
}
\textbf{Output}: $\{\breve{\mathbf{z}}_{i, 1} \oplus \breve{\mathbf{z}}_{i, 2} \oplus \cdots \oplus \breve{\mathbf{z}}_{i, M} | \forall i \in V\}$
}
\end{algorithm}

\section{Experiments}\label{sec:experiment}

\subsection{Experimental Setting}\label{sec:exp_set}

\textbf{Datasets} Cora, Citeseer, and Pubmed are three citation benchmark networks widely used in~\citep{Kipf2017SemiSupervisedCW,Hamilton2017InductiveRL,Velickovic2018GraphAN}, where nodes and edges denote documents and undirected citations respectively; each node is assigned with one topic and associated with bags-of-words features. We synthesize graphs with latent factors following~\citep{Ma2019DisentangledGC}. In detail, we first generate $m$ Erd{\H{o}}s-R{\'e}nyi random graphs with 1,000 nodes and 16 classes, where nodes connect each other with probability $p$ if they are in the same class, with probability $q$ otherwise. Then, we merge these generated graphs by summing the adjacency matrix and turning the element-value bigger than zero to one, to obtain the final synthetic graphs with $m$ latent factors. We set $q$ to 3e$^{-5}$ following~\citep{Ma2019DisentangledGC}, and tune $p$ value such that the average neighborhood size is between 39.5 and 40.5. Each node is initialized with the row of the adjacency matrix as the features and has $m$ labels. The data statistics are listed in the supplementary material.

\begin{table}[t]
\setlength{\abovecaptionskip}{2pt}\makeatletter\def\@captype{table}\makeatother\caption{Semi-supervised classification accuracies (\%)}
\label{tab:cla_acc}
\centering
\resizebox{0.42\textwidth}{!}{
\begin{tabular}{ccccc}
\hline
\multirow{2}{*}{\textbf{Method}} & \multirow{2}{*}{\textbf{Splits}} & \multicolumn{3}{c}{\textbf{Datasets}}               \\ \cline{3-5} 
                                 &                                          & \textbf{Cora} & \textbf{Citeseer} & \textbf{Pubmed} \\ \hline 
MLP                              & \multirow{5}{*}{Standard}                
&51.5$\pm$1.0   &46.5               &71.4     \\
MoNet                            &                                          
&82.2$\pm$0.7   &70.0$\pm$0.6       &77.7$\pm$0.6     \\
GCN                              &                                          
&81.9$\pm$0.8   &69.5$\pm$0.9       &79.0$\pm$0.5     \\
GAT                              &                                          
&82.5$\pm$0.5   &71.0$\pm$0.6       &77.0$\pm$1.3     \\
DisenGCN                         &                                          
&83.7           &73.4               &80.5     \\
LGD-GCN (ours)                          &                                          
&\textbf{84.9}$\pm$0.4   &\textbf{74.5}$\pm$0.8       &\textbf{81.3}$\pm$0.6     \\ \hline
MLP                              & \multirow{6}{*}{Random}                  
&58.2$\pm$2.1   &59.1$\pm$2.3       &70.0$\pm$2.1     \\
MoNet                            &                                          
&81.3$\pm$1.3   &71.2$\pm$2.0       &78.6$\pm$2.3     \\
GCN                              &                                          
&81.5$\pm$1.3   &71.9$\pm$1.9       &77.8$\pm$2.9     \\
GAT                              &                                          
&81.8$\pm$1.3   &71.4$\pm$1.9       &78.7$\pm$2.3     \\
DisenGCN                         &                                          
&81.4$\pm$1.6   &69.5$\pm$1.4       &79.1$\pm$2.3     \\
LGD-GCN (ours)                          &                                          
&\textbf{84.0}$\pm$1.3   &\textbf{72.0}$\pm$1.3       &\textbf{79.8}$\pm$2.3     \\ \hline
\end{tabular}
}
\end{table}
\begin{table}[t]
\centering
\setlength{\abovecaptionskip}{2pt}\makeatletter\def\@captype{table}\makeatother
\caption{Micro-F1 (Top) and Macro-F1 (Bottom) scores (\%) on synthetic graphs with different number of latent factors}
\label{tab:syn_mul_cal}
\resizebox{0.48\textwidth}{!}{
\begin{tabular}{cccccc}
\hline
                                          & \multicolumn{5}{c}{\textbf{Number of Latent Factors}} \\ \cline{2-6} 
\textbf{Method}           & 4        & 6        & 8        & 10        & 12       \\ \hline
MLP              
&79.3$\pm$0.5    &55.5$\pm$0.4        &37.0$\pm$0.8          &25.9$\pm$0.6        &21.2$\pm$0.8          \\
GCN                                        
&74.5$\pm$0.8    &56.3$\pm$0.7        &38.2$\pm$0.9          &28.0$\pm$0.7        &23.1$\pm$0.8          \\
DisenGCN                                   
&84.1$\pm$1.0   &60.4$\pm$0.9         &41.4$\pm$1.3          &29.4$\pm$0.7        &24.2$\pm$0.8          \\
LGD-GCN (ours)                                    
&\textbf{87.2$\pm$0.5}    &\textbf{65.0$\pm$0.5}        &\textbf{43.6$\pm$0.7}          &\textbf{30.2$\pm$0.5}           &\textbf{26.1$\pm$0.5}          \\\hline
MLP              
&77.9$\pm$0.7   &54.8$\pm$0.6        &36.0$\pm$0.8          &24.5$\pm$0.7       &20.1$\pm$0.9          \\
GCN                                        
&78.3$\pm$0.9    &55.6$\pm$0.9        &37.2$\pm$1.0          &26.9$\pm$0.5      &22.2$\pm$0.9          \\
DisenGCN                                   
&82.9$\pm$1.1    &59.9$\pm$1.0        &40.2$\pm$1.2          &28.1$\pm$0.7      &23.4$\pm$0.7          \\
LGD-GCN (ours)                                    
&\textbf{86.1$\pm$0.5}    &\textbf{64.2$\pm$0.6}        &\textbf{42.5$\pm$0.6}          &\textbf{28.8$\pm$0.5}           &\textbf{25.1$\pm$0.5}          \\ \hline
\end{tabular}
}
\end{table}

\textbf{Baseline Models.} We compare our model with several methods, including the state-of-the-art, as the baselines: MLP is a multi-layer perception; MoNet~\citep{Monti2017GeometricDL} is a mixture model CNN generalizing convolutional neural network to non-Euclidean graph data structure; GCN~\citep{Kipf2017SemiSupervisedCW} approximates graph Laplacian with Chebyshev expainsion; GAT~\citep{Velickovic2018GraphAN} combines the attention mechanism with graph neural networks to aggregate information with selective neighbors; DisenGCN attempts to learn disentangled node representations via a neighborhood routing mechanism.

\textbf{Hyper-parameters.} We set $d_{out}=64$ as the output dimension of each LGD-GCN's hidden layer and $T=7$ as the number of routing iterations, to follow GAT~\citep{Velickovic2018GraphAN} and DisenGCN respectively. For semi-supervised node classification on real-world datasets, we fix the number of channels $M$ as 4 for simplification. We use dropout $\sim\left[0, 1\right]$, learning rate $\sim\left[3\mathrm{e}{-3}, 1\right]$, weight decay $\sim\left[5\mathrm{e}{-5}, 0.2\right]$, update rate $\sim\left[0.1, 0.9\right]$ for $\mu_k$ and $\Sigma_k$, and the number of layers $\sim\{1, 2,...,10\}$. For multi-label classification on the synthetic datasets, with a slight difference, we fix dropout as 0.5, learning rate $\sim\left[5\mathrm{e}{-4}, 5\mathrm{e}{-3}\right]$, weight decay $\sim\left[1\mathrm{e}{-3}, 1\mathrm{e}{-2}\right]$, and $M\sim\{2,4,...,16\}$. 

Additionally, the regularization coefficients $\lambda_{space}$ and $\lambda_{div}$ as well as the density parameter $k$ are empirically searched from different ranges for different datasets as provided in the supplementary material. Then, we carefully tune the hyper-parameters defined above on the validation set using optuna~\citep{Akiba2019OptunaAN}. With the best hyper-parameters, we train the model in 1,000 epochs using the early-stopping strategy with a patience of 100 epochs, and report the average performance in 10 runs on the test split.

\subsection{Quantitative Evaluation}
In this section, we evaluate our model quantitatively in tasks of semi-supervised node classification and multi-label node classification.

\textbf{Semi-supervised Node Classification}.
In this task, we follow the experimental protocal established by~\citet{Kipf2017SemiSupervisedCW,Velickovic2018GraphAN}, and consider both standard split~\citep{Yang2016RevisitingSL} and random split. For random split, we uniformly sample the same number of instances as in the standard split in 10 times.

The results are listed in Table~\ref{tab:cla_acc} measured in classification accuracy. Since \citet{Shchur2018PitfallsOG} have conducted extensive evaluations in their work, we will quote their reported results for baseline methods. For DisenGCN, we not only collect their results, but also optimize and evaluate the model on the random splits using their source codes. For our model, considering the non-linear complexity of the real-world datasets, we adopt CkNN~\citep{berry2016consistent} in the module $\mathbb{LG}_{agg}$.

From the results, the proposed LGD-GCN consistently outperforms other baselines. Especially, our model is able to improve upon DisenGCN by a margin of 1.2\% and 2.6\% on Cora in standard and random splits, respectively. This demonstrates the benefits brought by absorbing rich and diverse global information. More importantly, real graphs are typically highly sparse as observed in Cora, Citeseer and Pubmed whose graphs contain an average neighbor number of 3.9, 2.8 and 4.5 for each node. In this case, our model is more effective in capturing long-range dependencies via the created shortcuts in the built geometric structures, which further explains the performance improvement.

\textbf{Multi-label Node Classification}. To further demonstrate our model's disentangling ability quantitatively, we apply MLP, GCN, DisenGCN, and our model to train graphs synthesized with various number of latent factors for multi-label node classification. Specifically, we randomly split each synthetic dataset into train/validation/test as 0.6/0.2/0.2, adopt kNN in the module $\mathbb{LG}_{agg}$, measure model performance in Micro-F1 and Macro-F1 scores, and report the results in Table~\ref{tab:syn_mul_cal}. It can be observed that our model consistently outperforms others while varying the number of latent factors, and especially achieves significant performance gains by (micro-f1) 4.6\% and (macro-f1) 4.3\% upon DisenGCN on the graph synthesized with six latent factors.

\begin{figure}[t]
    \centering
    \begin{subfigure}[b]{0.235\textwidth}
        \centering
        \includegraphics[width=0.9\textwidth]{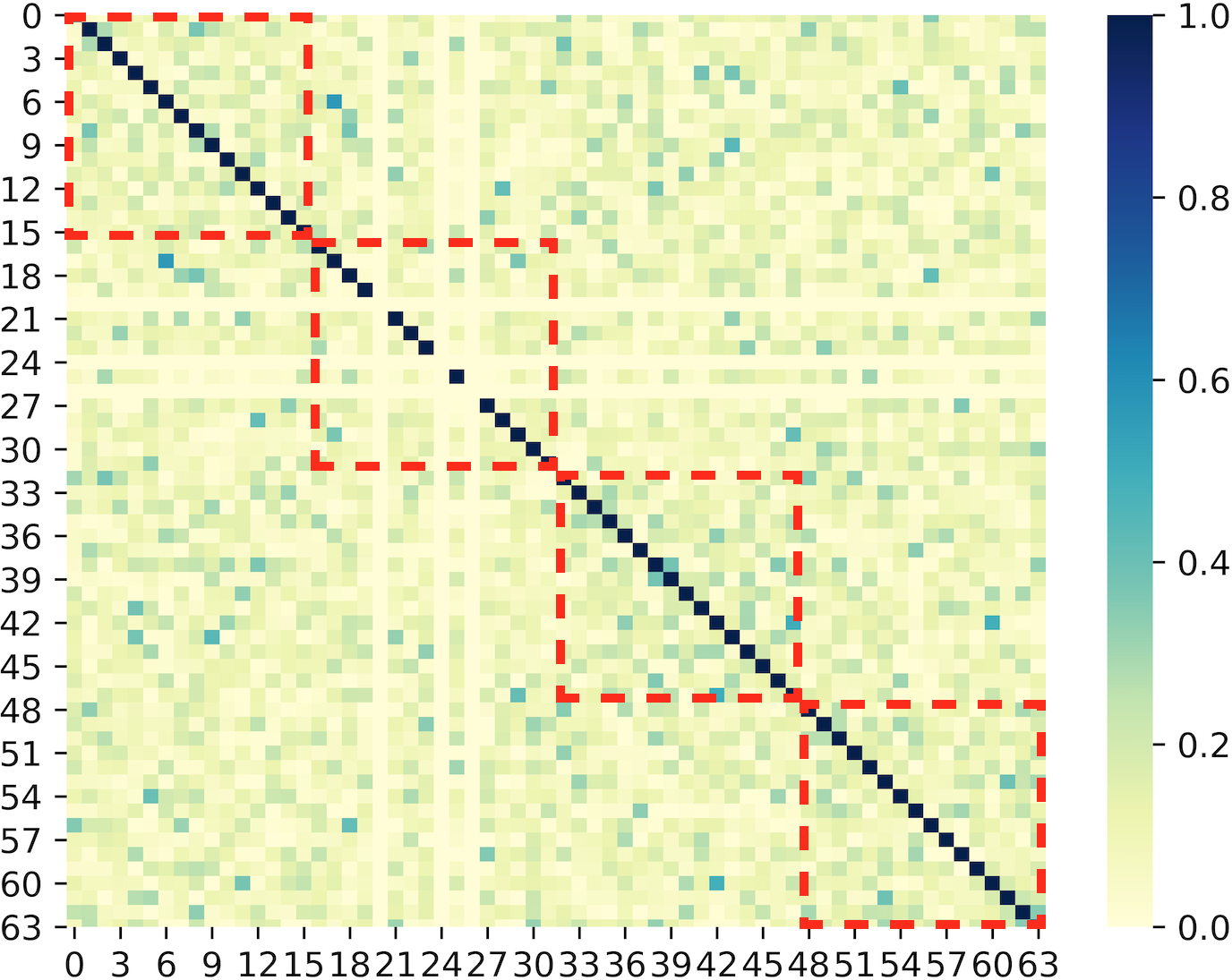}
        \caption{DisenGCN (1st Layer)}\label{fig:1L_disengcn}
    \end{subfigure}
    \hfill
    \begin{subfigure}[b]{0.235\textwidth}
        \centering
        \includegraphics[width=0.9\textwidth]{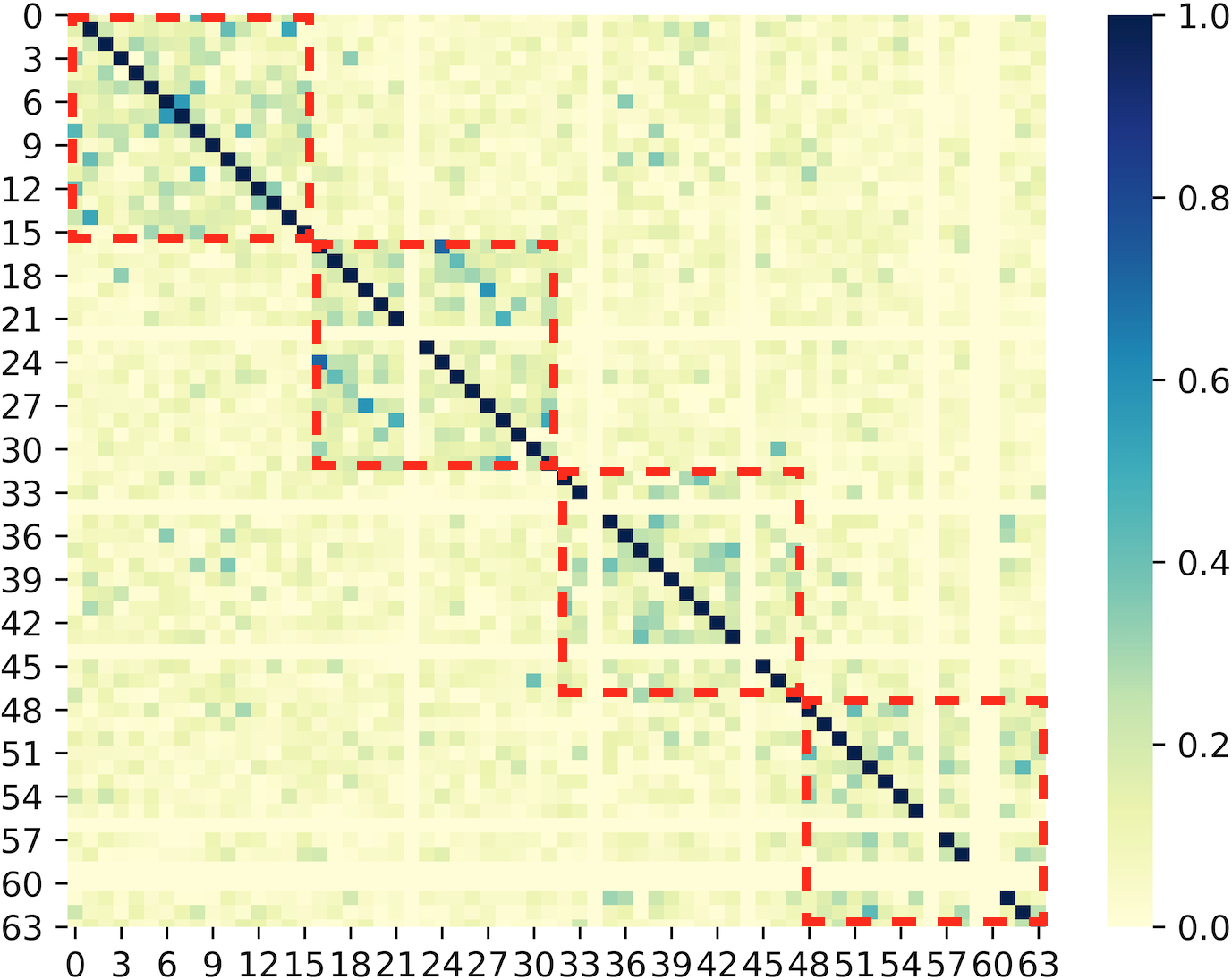}
        \caption{LGD-GCN (1st Layer)}\label{fig:1L_lgd2nd}
    \end{subfigure}
    \hfill
    \begin{subfigure}[b]{0.235\textwidth}
        \centering
        \includegraphics[width=0.9\textwidth]{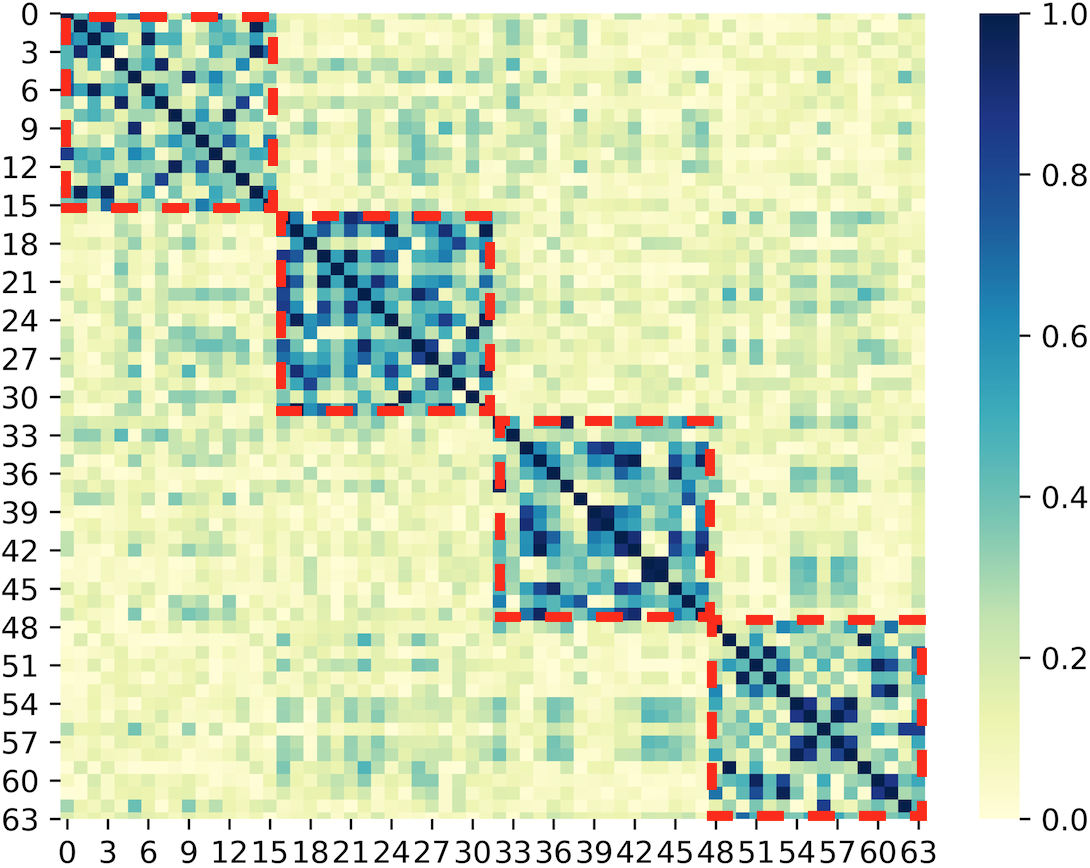}
        \caption{LGD-GCN (2nd Layer)}\label{fig:2L_lgd2nd}
    \end{subfigure}
    \hfill
    \begin{subfigure}[b]{0.235\textwidth}
        \centering
        \includegraphics[width=0.9\textwidth]{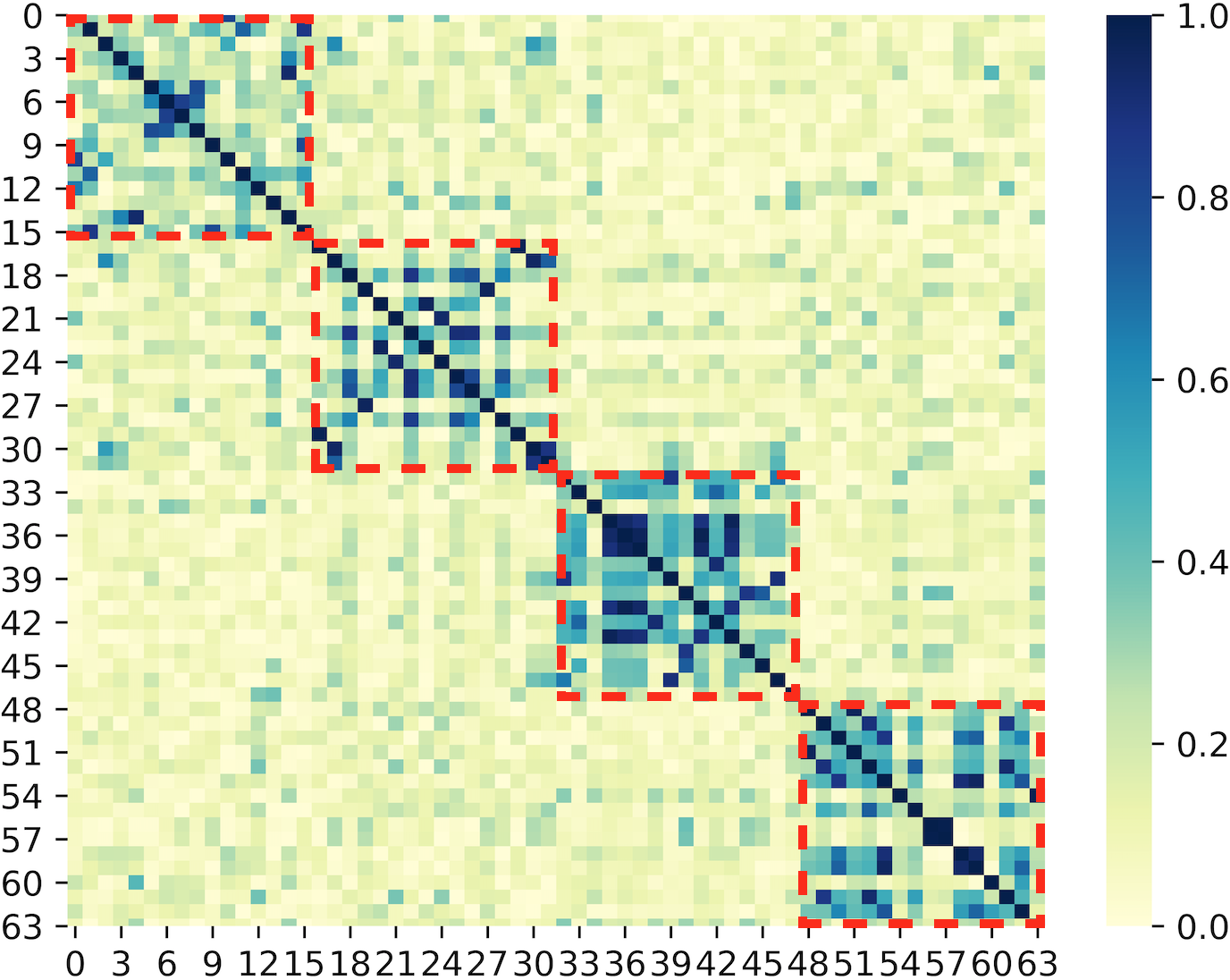}
        \caption{LGD-GCN$_{\mathbb{NG}}$ (2nd Layer)}\label{fig:2L_lgdng2nd}
    \end{subfigure}
    \caption{Features correlation analysis}
    \label{fig:FeatCorrelation}
\end{figure}

\subsection{Qualitative Evaluation}\label{sec:exp_q2}
The qualitative evaluation focuses on disentanglement performance and informativeness of learned embeddings. 

\textbf{Visualization of disentangled representations}. We give in Fig.~\ref{fig:lgd_good} a 2D visualization of the learned representations w.r.t. four latent factors on the synthetic graph. Compared to that of DisenGCN in Fig.~\ref{fig:disengcn_bad}, our model displays a highly disentangled pattern with consistent and diverse latent factors, evidenced by the intra-factor compactness and inter-factor separability; it also indicates the nodes carry the common type of factor-specific global information.

\textbf{Correlation of disentangled features}. The correlation analysis of the latent features, learned by DisenGCN and our model on test split of the graph synthesized with four latent factors, is presented in Fig.~\ref{fig:FeatCorrelation}. As observed, our model showcases a more block-wise correlation pattern, which becomes denser in the second layer, indicating the enhanced interpretability. We also analyze the feature correlation of our model while ablating the module $\mathbb{LG}_{agg}$, denoted as LGD-GCN$_{\mathbb{NG}}$ in Fig.~\ref{fig:2L_lgdng2nd}. Though the block-wise pattern in Fig.~\ref{fig:2L_lgdng2nd}  can still be observed, it is obviously weaker than that of LGD-GCN in Fig.~\ref{fig:2L_lgd2nd}. This verifies the significance of $\mathbb{LG}_{agg}$; the captured factor-specific global information strengthens the factor-aware feature correlation, and enhances the interpretability  of the learned representations.

\begin{figure}[t]
    \centering
    \begin{subfigure}[b]{0.235\textwidth}
        \centering
        \includegraphics[width=0.9\textwidth]{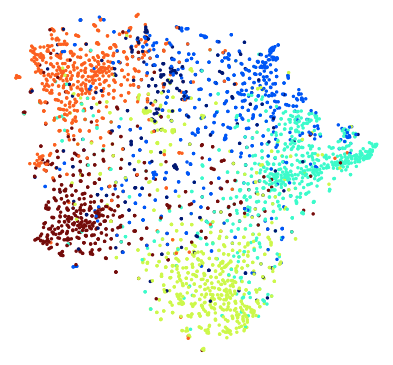}
        \caption{\small{DisenGCN}}\label{fig:cite_disengcn}
    \end{subfigure}
    \hfill
    \begin{subfigure}[b]{0.235\textwidth}
        \centering
        \includegraphics[width=0.9\textwidth]{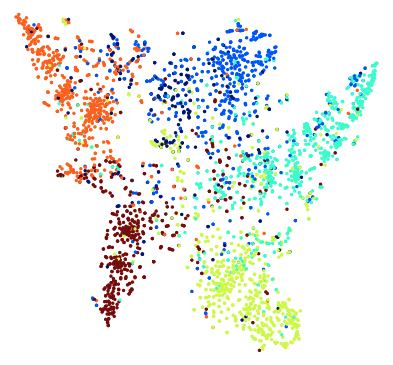}
        \caption{\small{LGD-GCN}}\label{fig:cite_lgd}
    \end{subfigure}     
\caption{Visualization of node embedding on Citeseer}\label{fig:CiteVisual}
\end{figure}

\textbf{Visualization of node embeddings}. Fig.~\ref{fig:CiteVisual} provides a intuitive comparison between the learned node embeddings of DisenGCN and our model on Citeseer dataset. It can be observed that the proposed LGD-GCN learns better node embeddings and shows a high inter-class similarity and intra-class difference. This is because our model learns more informative node aspects by absorbing rich factor-specific global information, leading to increasing discriminative power.

\subsection{Parameter and ablation analysis}
We investigate the sensitivity of hyper-parameters, and perform ablation analysis over the proposed modules on real-world and synthetic datasets.

\begin{figure}[t]
     \centering
     \begin{subfigure}[b]{0.235\textwidth}
         \centering
         \includegraphics[width=1\columnwidth]{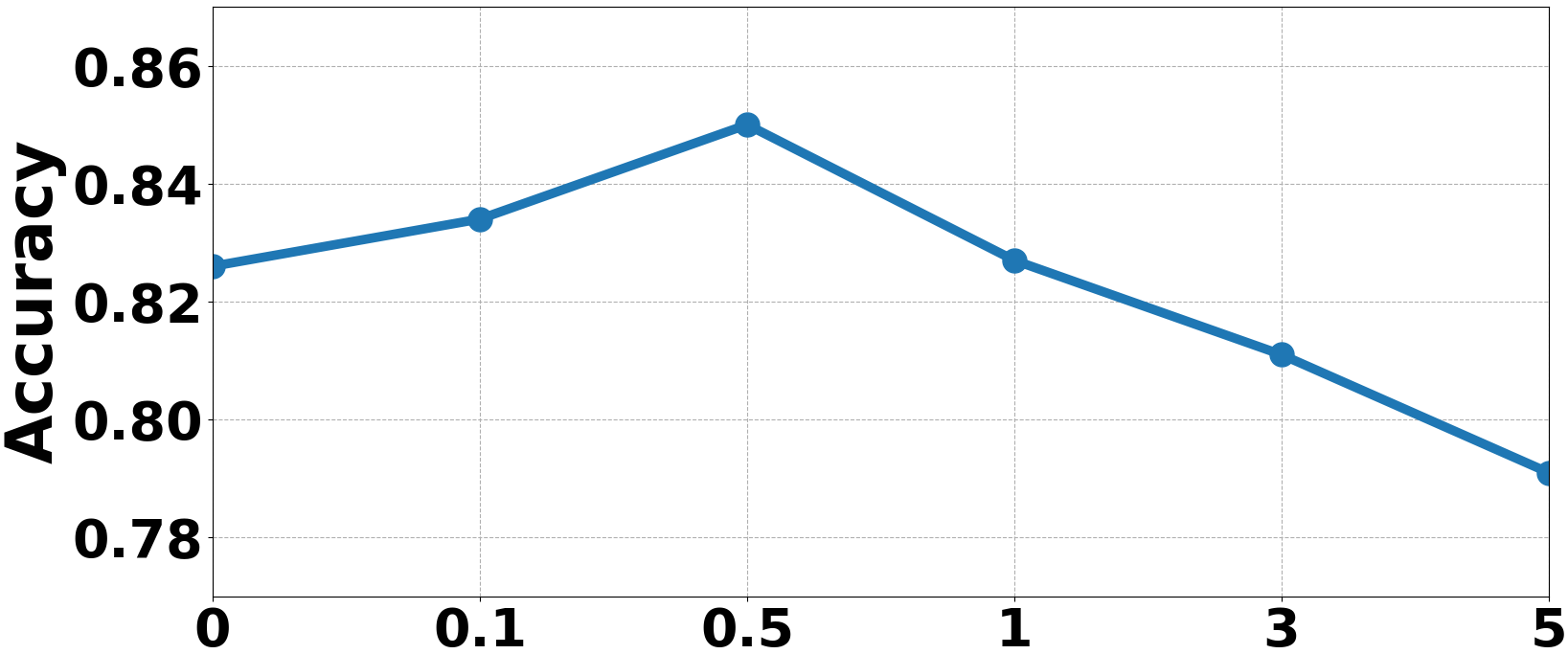}
         \caption{\small{Cora}}
         \label{fig:consensi_cora}
     \end{subfigure}
     \hfill
     \begin{subfigure}[b]{0.235\textwidth}
         \centering
         \includegraphics[width=1\columnwidth]{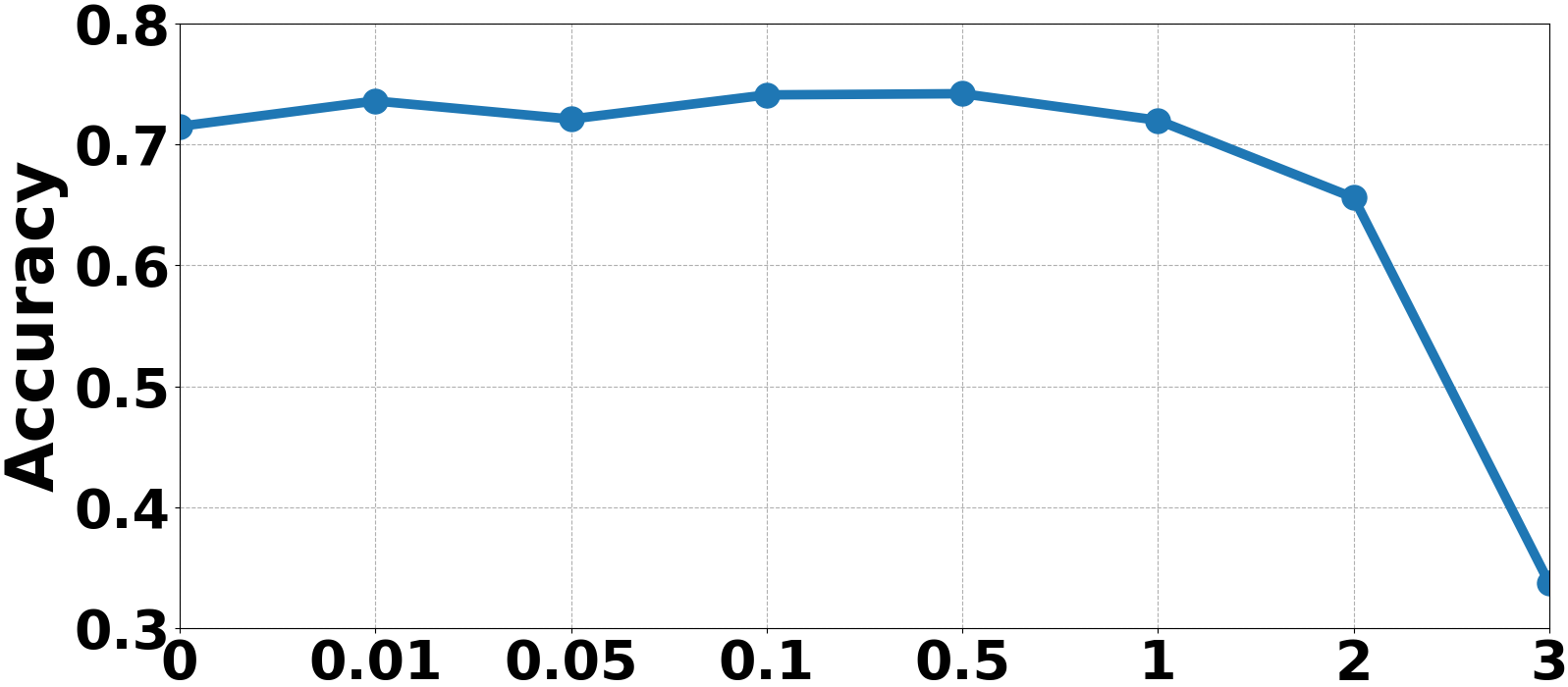}
         \caption{\small{Citeseer}}
         \label{fig:consensi_cite}
     \end{subfigure}
\caption{Analysis of parameter $\lambda_{space}$}
\label{fig:ConSensi}
\end{figure}

\textbf{Analysis of consistency coefficient $\lambda_{space}$}. We plot the learning performance of our model w/o $\mathcal{L}_{div}$  while varying $\lambda_{space}$ in Eq.~(\ref{eq:ttloss}) e.g. from 0 to 5 on Cora in Fig.~\ref{fig:consensi_cora}. The accuracy goes up first and drops slowly. Practically, promising performance can be attained on Cora by choosing $\lambda_{space}$ from [0.1, 1].

\begin{figure}[t]
    \centering
    \begin{subfigure}[b]{0.235\textwidth}
         \centering
         \includegraphics[width=1\columnwidth]{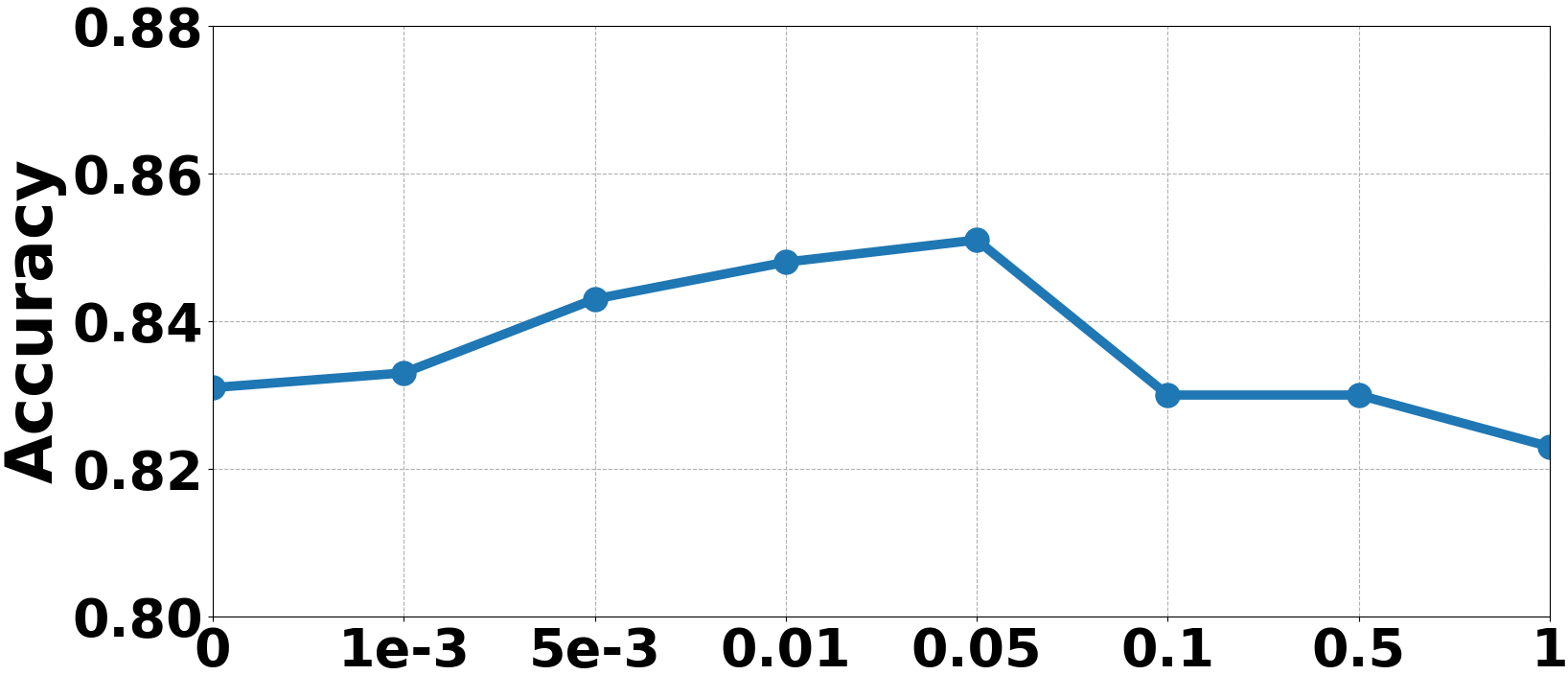}
         \caption{\small{Cora}}
         \label{fig:divsensi_cora}
     \end{subfigure}
     \hfill
     \begin{subfigure}[b]{0.235\textwidth}
         \centering
         \includegraphics[width=1\columnwidth]{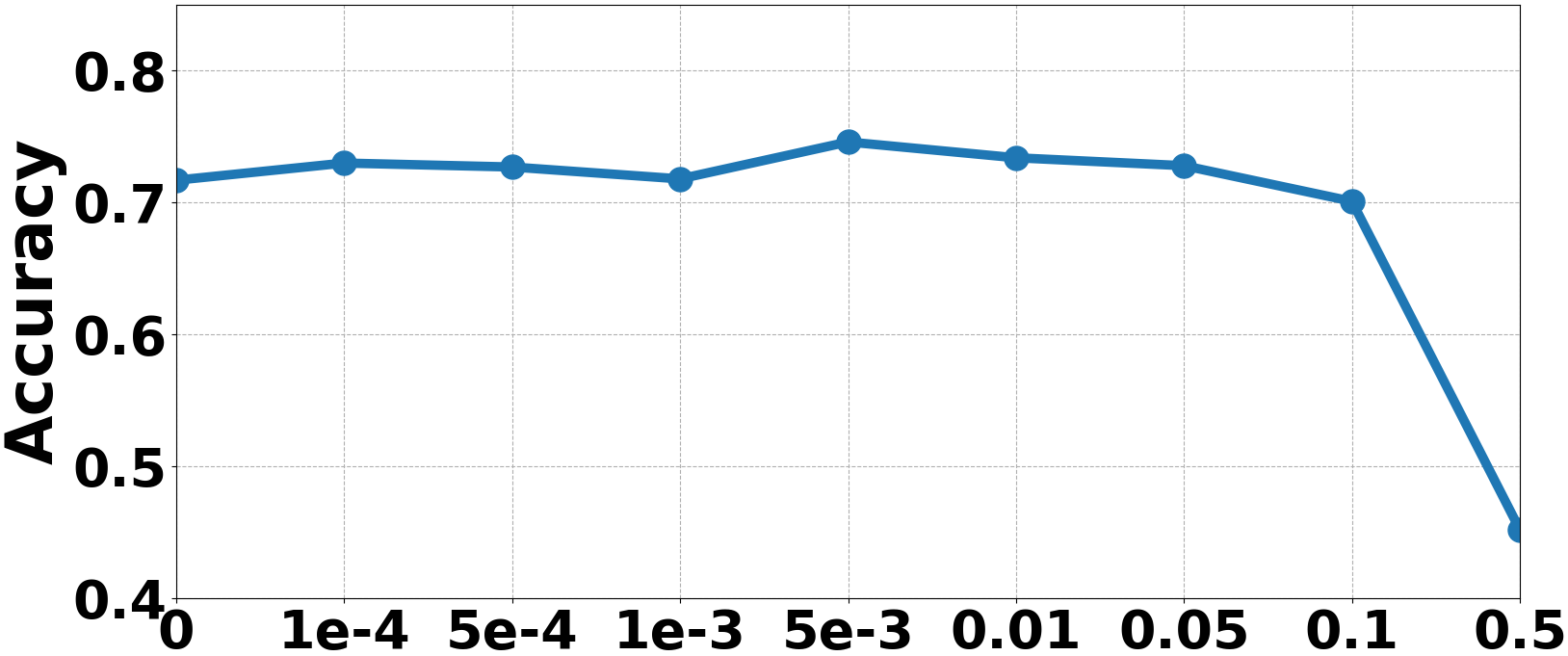}
         \caption{\small{Citeseer}}
         \label{fig:divsensi_cite}
     \end{subfigure}
\caption{Analysis of parameter $\lambda_{div}$}
\label{fig:DivSensi}
\end{figure}

\textbf{Analysis of diversity coefficient $\lambda_{div}$}. We then test the effect of $\lambda_{div}$ in Eq.~(\ref{eq:ttloss}), and vary it from e.g. 0 to 0.5 on Citeseer. $\lambda_{div}$ is relatively robust within a certain range e.g. [0, 0.1] for Citeseer in Fig.~\ref{fig:divsensi_cite}. Once out of that range, the results drops to a low point, suggesting overly focusing on diversity is harmful to model performance.

\begin{figure}[t]
     \centering
     \begin{subfigure}[b]{0.235\textwidth}
         \centering
         \includegraphics[width=1\columnwidth]{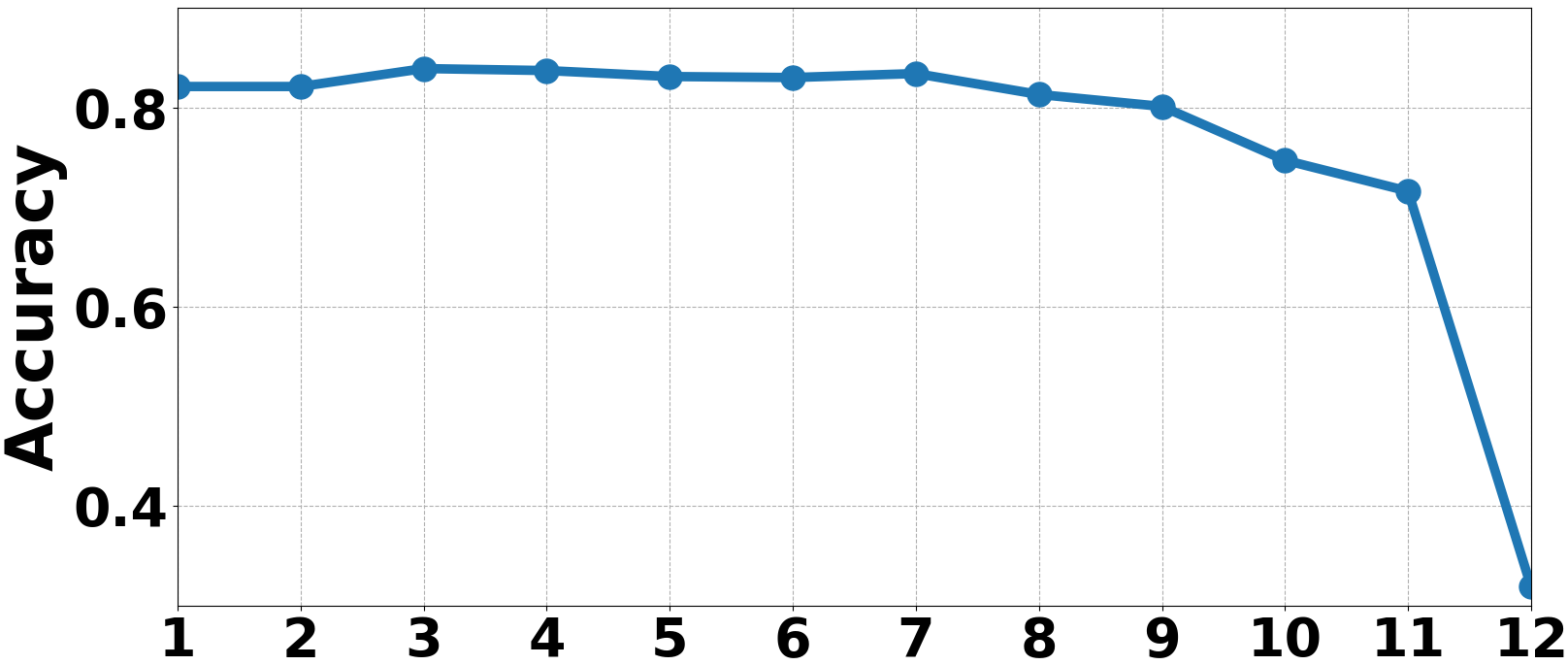}
         \caption{\small{Cora}}
         \label{fig:ksensi_cora}
     \end{subfigure}
     \hfill
     \begin{subfigure}[b]{0.235\textwidth}
         \centering
         \includegraphics[width=1.0\columnwidth]{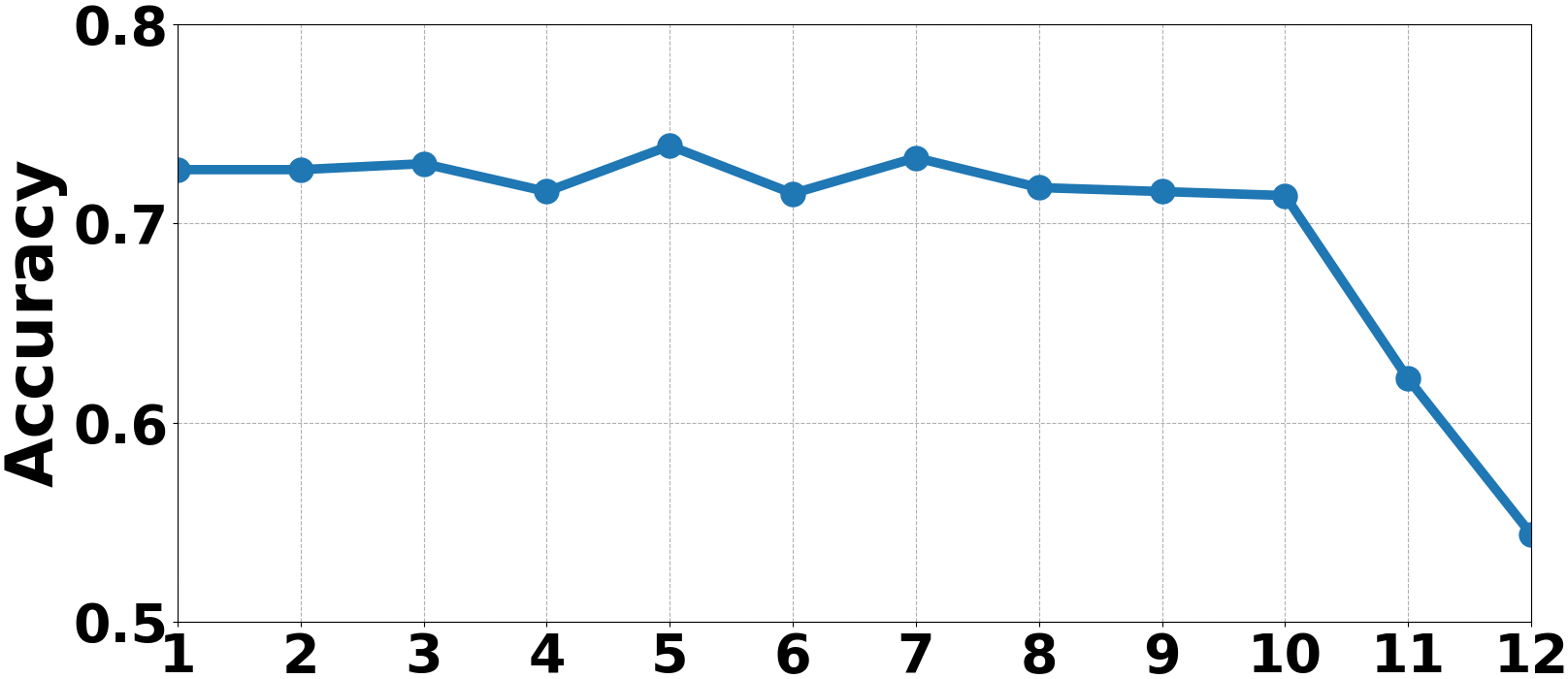}
         \caption{\small{Citeseer}}
         \label{fig:ksensi_cite}
     \end{subfigure}
\caption{Analysis of parameter $k$}
\label{fig:KSensi}
\end{figure}

\textbf{Analysis of density parameter $k$}. Fig.~\ref{fig:KSensi} displays the impact of $k$ from 1 to 12 on Cora and Citeseer. The results are relatively stable while selecting $k$ from a wide range, e.g. 1 to 8 on Cora and 1 to 10 on Citeseer. However, as $k$ getting larger, the accuracy performance deteriorates obviously. It probably because larger $k$ may introduce noisy edges, leading to inappropriate information sharing.

\begin{figure}[t]
    \captionsetup[subfigure]{labelformat=empty}
    \centering
    \begin{subfigure}[b]{1.0\columnwidth}
    \centering
    \includegraphics[width=0.705\textwidth]{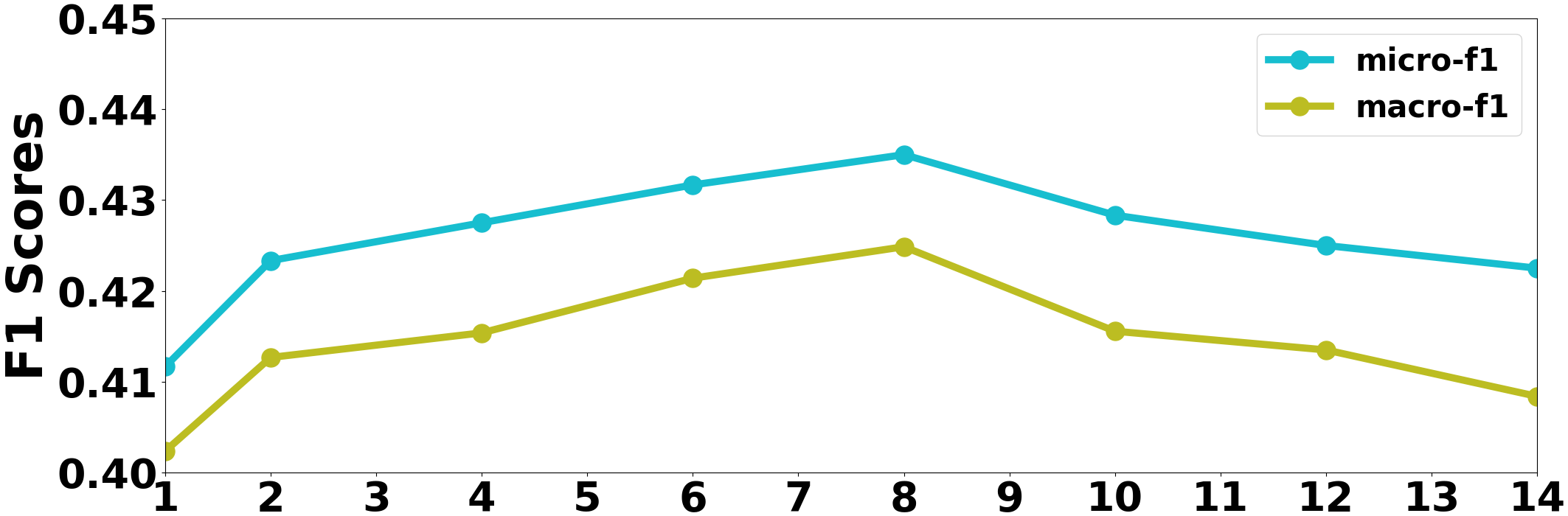}
    \caption{\small{Graph synthesized with eight latent factors}}
    \end{subfigure}
\caption{Analysis of parameter $M$}
\label{fig:channel_sensi}
\end{figure}

\textbf{Analysis of the number of channels $M$}. We study the influence of the number of channels $M$ on the synthetic graphs generated with eight latent factors. From Fig.~\ref{fig:channel_sensi}, our model performs the best when the number of channels is around eight, the true number of the latent factors.


\textbf{Ablation analysis}. We validate the contributions of the proposed modules denoted by $\mathcal{L}_{space}$, $\mathcal{L}_{div}$, and $\mathbb{LG}_{agg}$ in  node classification. From Table~{\ref{tab:abla_study}}, we can see that both modules can independently and jointly improve the accuracy.

\begin{table}[ht]
\centering
\setlength{\abovecaptionskip}{2pt}\makeatletter\def\@captype{table}\makeatother\caption{Ablation analysis in node classification accuracies.}
\label{tab:abla_study}
\resizebox{0.47\textwidth}{!}{
\begin{tabular}{ccccc}
\hline
\textbf{Components} & \textbf{}                     & \multicolumn{1}{c}{\textbf{Cora}} & \textbf{Citeseer} & \textbf{Pubmed} \\ \hline
-& \multirow{6}{*}{\textbf{\%}} 
&81.9$\pm$1.1                               &70.4$\pm$1.6                   &78.9$\pm$0.7                \\
$\mathcal{L}_{space}$&                               
&83.0$\pm$0.5                               &72.4$\pm$1.3                   &79.1$\pm$0.8                 \\
$\mathcal{L}_{space}$+$\mathcal{L}_{div}$&
&83.6$\pm$0.6                               &72.4$\pm$1.1                   &79.1$\pm$0.4                 \\
$\mathcal{L}_{space}$+$\mathbb{LG}_{agg}$&
&84.4$\pm$0.3                               &74.0$\pm0.7$           &\textbf{81.3}$\pm$0.6                 \\
$\mathcal{L}_{space}$+$\mathcal{L}_{div}$+$\mathbb{LG}_{agg}$&
&\textbf{84.9}$\pm$0.4                      &\textbf{74.5}$\pm$0.8          &81.2$\pm$0.7           
  \\ \hline
\end{tabular}}
\end{table}

\section{Conclusion}
In this paper, we propose a novel framework, termed Local and Global Disentangled Graph Convolutional Network (LGD-GCN), to disentangle node representations with strengthened intra-factor consistency and promoted inter-factor diversity.
Extensive experiments demonstrate the improved performance in node classification and enhanced interpretability of the proposed LGD-GCN over existing state-of-the-art approaches.

\bibliography{LGD}

\clearpage
\appendix
\section{Supplementary Material}
In the supplementary material, for reproducibility, we provide the dataset information, algorithm and optimization procedure, and hyper-parameters' searching ranges. Finally, we show more visualization results to validate our conclusion in the manuscript.

\subsection{Dataset Statistics}
We list the information of datasets evaluated in the manuscript in Table~\ref{tab:realdat_sta} and Table~\ref{tab:syndata_sta}. For real-world datasets, we only use 20 labeled nodes per class but with all the rest nodes unlabeled for training, another 500 nodes for validation and early-stopping, and 1,000 nodes from the rest for testing. For synthetic datasets, we specify the parameters including probability $p$ and probability $q$ for generating synthetic graphs with various number of latent factors.

\subsection{Algorithm and Optimization}
Algorithm~\ref{alg:lgd_opt} illustrates the optimization procedures in pseudo-codes.


\subsection{Additional Results}
\begin{figure}[ht]
    \captionsetup[subfigure]{labelformat=empty}
    \centering
    \begin{subfigure}[b]{0.28\textwidth}
        \centering
        \includegraphics[width=.85\textwidth]{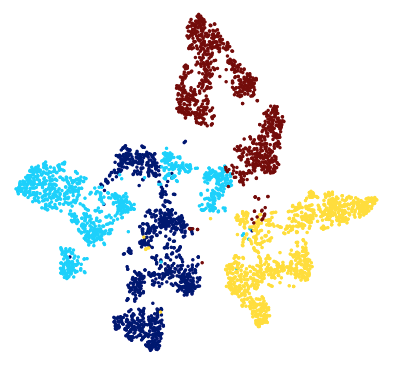}
        \caption{LGD-GCN w/o $\mathbb{LG}_{agg}$}
    \end{subfigure}
    \caption{Visualization of disentangled representations on a synthetic graph with four latent factors}\label{fig:nglgd_not_bad}
\end{figure}

\begin{figure}[ht]
    \begin{subfigure}[b]{0.23\textwidth}
         \centering
         \includegraphics[width=.85\textwidth]{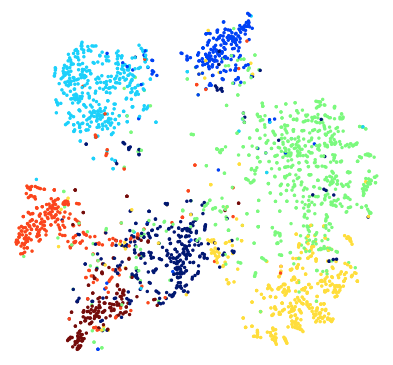}
         \caption{DisenGCN}
         \label{fig:cora_disengcn}
    \end{subfigure}
    \hfill
    \begin{subfigure}[b]{0.23\textwidth}
         \centering
         \includegraphics[width=.85\textwidth]{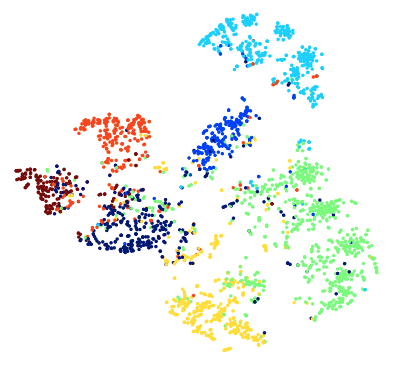}
         \caption{LGD-GCN}
         \label{fig:cora_lgd}
    \end{subfigure}
    \caption{Visualization of node embedding on Cora}
    \label{fig:CoraVisual}
\end{figure}

To further verify the importance of the module $\mathbb{LG}_{agg}$, we ablate it from our model and visualize the disentangled representations on the synthetic graph with four latent factors in Fig.~\ref{fig:nglgd_not_bad}. Comparing to that of the original model, we can witness an evident performance drop by the weakened intra-factor compactness. Even worse, the integrated blue set in Fig.~\ref{fig:lgd_good} is broken into two disjoint clusters in Fig.~\ref{fig:nglgd_not_bad} by turning off the module $\mathbb{LG}_{agg}$, indicating its effectiveness. We also show the visualization of node embedding learned on Cora dataset in Fig.~\ref{fig:CoraVisual}. Similar to that on Citeseer dataset in Fig.~\ref{fig:CiteVisual}, our model learns better embeddings, evidenced by intra-class compactness and inter-class separability.

\begin{table}[ht]
\centering 
\setlength{\abovecaptionskip}{2pt}\makeatletter\def\@captype{table}\makeatother
\caption{Real-world Dataset Statistics}\label{tab:realdat_sta}
\resizebox{0.39\textwidth}{!}{
\begin{tabular}{cccc}
\hline
\textbf{Dataset} & \textbf{Cora} & \textbf{Citeseer} & \textbf{Pubmed} \\ \hline
Nodes            & 2708          & 3327              & 19717           \\
Avg-Neighbors    & 3.9           & 2.8               & 4.5                \\
Features         & 1433          & 3703              & 500             \\
Classes          & 7             & 6                 & 3               \\ \hline
Train            & 140           & 120               & 60              \\
Validation       & 500           & 500               & 500             \\
Test             & 1000          & 1000              & 1000            \\
\hline
\end{tabular}}
\end{table}

\begin{table}[t]
\centering
\setlength{\abovecaptionskip}{2pt}\makeatletter\def\@captype{table}\makeatother
\caption{Parameters for generating synthetic datasets}\label{tab:syndata_sta}
\resizebox{0.47\textwidth}{!}{
\begin{tabular}{cccccc}
\hline
\textbf{\# Latent Factors} & \textbf{4} & \textbf{6} & \textbf{8} & \textbf{10} & \textbf{12} \\ \hline
probablity $p$                      & 0.164      & 0.110       & 0.082      & 0.065       & 0.055       \\
probability $q$                     & 3e-5       & 3e-5       & 3e-5       & 3e-5        & 3e-5        \\ \hline
\end{tabular}}
\end{table}

\begin{algorithm}[t]
\caption{\small{LGD-GCN's Optimization Procedure}}\label{alg:lgd_opt}
\SetAlgoLined
\small{
\textbf{Input}: $\{\mathbf{h}_i^{(0)} \in \mathbb{R}^{d_{in}} | \forall i \in V\}$; $l_r$ be the learning rate; $U_r$ be the update rate for $\boldsymbol{\mu}_m^{(l)} \in \mathbb{R}^\frac{d_{out}}{M}$ and $\boldsymbol{\Sigma}_m^{(l)} \in \mathbb{R}^{\frac{d_{out}}{M} \times \frac{d_{out}}{M}}$, $\forall m \in \{1,2,...,M\}, \forall l=1,2,..,L$, $\mathbf{F}_{\Theta}$ refers to the proposed LGD-GCN with learnable weights $\Theta$. \\

\For{$\text{number of training epochs}$}{
Compute $\mathcal{L}_{total}$ by Eq.~(\ref{eq:ttloss}); Update $\Theta$ using Adam optimizer~\citep{Kingma2015AdamAM} with learning rate $l_r$. \\
    \For{$l=1,2,...,L$}{
        \For{$i \in V$}{
        $\{\breve{\mathbf{z}}_{i,1}^{(l)},\breve{\mathbf{z}}_{i,2}^{(l)},...,\breve{\mathbf{z}}_{i,M}^{(l)}\} \gets \mathbf{F}_{\Theta}(\breve{\mathbf{h}}_i^{(l-1)})$\;
        }
        \For{$m = 1,2,...M$}{
            $\accentset{\ast}{\boldsymbol{\mu}}_{m}^{(l)} \gets \frac{1}{|V|}\sum_{i \in V}\breve{\mathbf{z}}_{i,m}$\;
            $\accentset{\ast}{{\boldsymbol{\Sigma}}}_{m}^{(l)} \gets \frac{1}{|V|}\sum_{i \in V}(\breve{\mathbf{z}}_{i, m} - \boldsymbol{\mu}_m^{(l)})(\breve{\mathbf{z}}_{i, m} - \boldsymbol{\mu}_m^{(l)})^T$\;
            $\boldsymbol{\mu}_m^{(l)} \gets (1-U_r)\boldsymbol{\mu}_{m}^{(l)} + U_r\accentset{\ast}{\boldsymbol{\mu}}_m^{(l)}$\;
            $\boldsymbol{\Sigma}_m^{(l)} \gets (1-U_r)\boldsymbol{\Sigma}_{m}^{(l)} + U_r\accentset{\ast}{\boldsymbol{\Sigma}}_m^{(l)}$\;
        }
    }
}
}
\end{algorithm}

\end{document}